\documentclass[letterpaper, 10 pt, conference]{ieeeconf}  

\IEEEoverridecommandlockouts                              

\usepackage[utf8]{inputenc}
\usepackage[T1]{fontenc}
\usepackage{textcomp}
\usepackage{graphicx} 

\usepackage{pgfplots}
\pgfplotsset{compat=newest}
\usepackage{subfig}
\usepackage{multirow}
\usepackage{tabularx}
\usepackage[hidelinks]{hyperref}
\usepackage{array}
\usepackage{float}
\usepackage{graphbox}
\usepackage{tikz,tikz-3dplot}
\usetikzlibrary{arrows,arrows.meta,automata,backgrounds,calc,chains,%
decorations.markings,decorations.pathreplacing,decorations.pathmorphing,%
matrix,positioning,shapes,shapes.geometric,shapes.symbols,spy,trees,tikzmark}
\usepackage{balance}
\usepackage{bm}
\usepackage[binary-units=true,product-units=single,per-mode=symbol,range-units=single,detect-all]{siunitx}
\DeclareSIUnit\pixel{px}
\usepackage{amsmath} 
\usepackage{amssymb}  
\usepackage{acronym}
\usepackage{tablefootnote}

\newcommand{\tcb}[1]{\textcolor{blue}{#1}}
\newcommand{\med}[1]{\text{median}\left(#1\right)}
\newcommand{\norm}[1]{\left\lVert#1\right\rVert}
\newcommand{\abs}[1]{\lvert#1\rvert}
\DeclareMathOperator*{\argmin}{arg\,min}
\DeclareMathOperator*{\argmax}{arg\,max}

\newcommand{\reffig}[1]{Fig.~\ref{#1}}
\newcommand{\reftab}[1]{Tab.~\ref{#1}}
\newcommand{\refsec}[1]{Sec.~\ref{#1}}
\newcommand{\refeq}[1]{Eq.~\ref{#1}}
\newcommand{\etal}{et al.~}
\newcommand{\citep}[1]{(\cite{#1})}

\newcommand{\wrt}{~w.r.t.~}

\newcommand{\eg}{e.g.,\ }

\newacro{ICP}{Iterative Closest Point}
\newacro{GICP}{Generalized ICP}
\newacro{NDT}{Normal distribution transform}
\newacro{mocap}[MoCap]{Motion Capture}
\newacro{uav}[UAV]{unmanned aerial vehicle}
\newacro{EM}{Expectation Maximization}
\newacro{gmm}[GMM]{Gaussian Mixture Model}
\newacro{deque}{double-ended queue}
\newacro{voxel}{volume element}
\newacro{surfel}{surface element}
\newacro{UGVs}{unmanned ground vehicles}
\newacro{UAVs}{unmanned aerial vehicles}
\newacro{SLAM}{Simultaneous Localization and Mapping}
\newacro{MARS}{Multi-Adaptive-Resolution-Surfel}
\newacro{ATE}{root mean squared absolute trajectory error}

\usepackage[font=small]{caption}

\makeatletter
\tikzset{
from/.style args={#1 to #2}{
        above right={0cm of #1},
        /utils/exec=\pgfpointdiff
            {\tikz@scan@one@point\pgfutil@firstofone(#1)\relax}
            {\tikz@scan@one@point\pgfutil@firstofone(#2)\relax},
        minimum width/.expanded=\the\pgf@x,
        minimum height/.expanded=\the\pgf@y}}
\makeatother

\title{\LARGE \bf
Real-time Multi-Adaptive-Resolution-Surfel 6D LiDAR Odometry\\ using Continuous-time Trajectory Optimization 
}

\author{Jan Quenzel and Sven Behnke%
\thanks{This work has been supported by the German Federal Ministry of Education and Research (BMBF) in the project ``Kompetenzzentrum: Aufbau des Deutschen Rettungsrobotik-Zentrums (A-DRZ)''}%
\thanks{Institute for Computer Science VI, Autonomous Intelligent Systems, University of Bonn, Friedrich-Hirzebruch-Allee 8, 53115 Bonn, Germany,
		{\tt\small quenzel@ais.uni-bonn.de}%
}
}

\begin{document}

\maketitle
\thispagestyle{empty}
\pagestyle{empty}

\begin{abstract}
\ac{SLAM} is an essential capability for autonomous robots, but due to high data rates of 3D LiDARs real-time SLAM is challenging. 
We propose a real-time method for 6D LiDAR odometry. Our approach combines a continuous-time B-Spline trajectory representation with a \ac{gmm} formulation to jointly align local multi-resolution surfel maps. Sparse voxel grids and permutohedral lattices ensure fast access to map surfels, and an adaptive resolution selection scheme effectively speeds up registration.
A thorough experimental evaluation shows the performance of our approach on multiple datasets and during real-robot experiments.
\end{abstract}

\section{Introduction}

LiDAR plays a major role in environment perception and mapping for autonomous driving~\cite{geiger2012cvpr}, \ac{UGVs}~\cite{shan2018legoloam} and \ac{UAVs}~\cite{beul2018iros}. \ac{SLAM} and odometry systems provide the basis for many autonomy or assistance functionalities like out-of-sight operation in GNSS-denied environments and reduce strain on the operator while improving their awareness of the surrounding area.

Despite much progress, robustness and reliability in crowded, dynamic scenes and close to structures remain challenging. In recent years, the amount and density of LiDAR measurements increased tremendously which poses new challenges for real-time processing of large point clouds. These factors are essential when a risk-minimizing state, like stopping, is difficult to maintain, \eg for \ac{UAVs}.

Most odometry and SLAM systems do not take full advantage of dependencies between consecutive LiDAR scans when aligning a scan against a local map or a previous scan, only jump-starting their registration with prior motion estimates. This may lead to unrealistic jumps in the trajectory since the sensor motion imposes a dependence between consecutive scans. We address this limitation with a continuous-time trajectory representation.

\begin{figure}
  \centering
  \resizebox{1.0\linewidth}{!}{\input{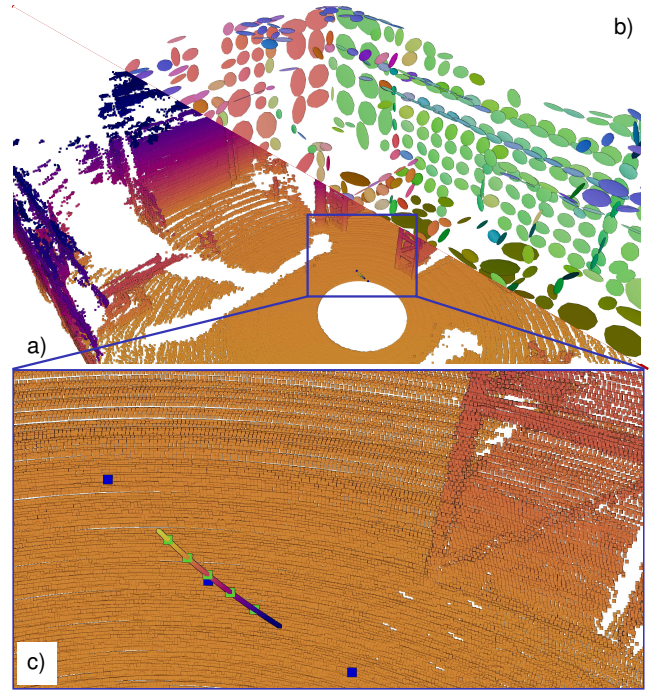}}
  \caption{Multi-resolution surfel maps with adapted resolution represent the point clouds a) during registration with a continuous-time trajectory B-spline. The surfel b) color depends on their normal, while point color depends on height. c) The control points (blue dots) interpolate the scan poses (green dots) and form the spline (blue to yellow).}
  \label{fig:teaser}
\end{figure}

The main contribution of this paper is our novel \ac{MARS} LiDAR odometry system\footnote{We will open-source our LiDAR odometry at: \url{https://github.com/AIS-Bonn/lidar_mars_registration}. An accompanying video is available at \url{https://ais.uni-bonn.de/videos/iros2021_quenzel/}} that jointly registers multiple point clouds against a local multi-resolution surfel map using a continuous-time Lie group B-Spline~\cite{sommer2020cvpr}, as visualized in \reffig{fig:teaser}. We adaptively select the most efficient resolution for registration and employ (block-)sparse voxel grids or permutohedral lattices for surfel map storage. We modify the \ac{gmm} formulation of Droeschel \etal\cite{david2017ras} to improve numerical stability and introduce a normal-distance-based weighting.
We thoroughly evaluate our system to support our key claims, which are:
\begin{itemize}
\item our system provides reliable pose estimates with state-of-the-art quality on a variety of datasets,
\item our \ac{gmm} is more numerically stable and more approriate given the typical LiDAR sensor geometry,
\item our adaptive resolution selection effectively reduces the required computation without degrading accuracy,
\item our system runs in real-time onboard a UAV, enabling safe operation in GNSS-denied environments.
\end{itemize}

\section{Related Work}
Point cloud registration is a well-researched topic and has wide applicability~\cite{HolzITRB15}. A basic registration method is the \ac{ICP}~\cite{besl1992icp} algorithm. \ac{ICP} aligns scan and model point clouds in an iterative two-step process. In the first step, the algorithm establishes correspondences between the two point clouds. The second step calculates a transformation to reduce the distance between all corresponding points and repeats both steps with the transformed scan until reaching some optimization criteria. The original formulation assumes perfect correspondences, thus suffers in reality when the sampling locations differ due to a moving sensor. 

Another popular scan registration approach is the \ac{NDT}~\cite{magnusson2007ndt}. Here, the model point cloud is represented by normal distributions within a regular grid. 
This reduces the memory consumption as well as computation time for nearest-neighbor searches. \ac{NDT} aims at maximizing the likelihood of all scan points to observe the underlying \ac{surfel} described by the normal distributions.

Similarly, Segal \etal\cite{segal2009gicp} rephrased \ac{ICP} within a probabilistic framework that allows incorporating information from the correspondence's covariance. Hence, points on a planar surface will be pulled together in normal direction but have more leeway along the surface.

Once there is a method to register (feature-)points, continuously estimating the sensor pose \wrt an updating local map becomes possible.
The localization and mapping (LOAM) approach~\cite{zhang2014loam} extracts feature points on planar surfaces and edges from the current scan based on the local curvature. Matching features against the previous scan allows estimation of the relative motion at scan frequency. Previous pose estimates help to undistort a newly incoming scan. Every $n$-th scan is then further processed in the mapping thread and aligned against and integrated into the map. LeGO-LOAM~\cite{shan2018legoloam} adapted the general approach for horizontally placed LiDARs on UGVs under the assumption of always being able to measure the ground plane. More recently, filter-based approaches \cite{qin2020lins,ye2019liom} use the same feature extraction to directly fuse LiDAR with IMU.

SuMa~\cite{behley2018rss} performs a projection-based data association to avoid the need for costly nearest neighbor associations. For this, it projects the current point cloud from spherical coordinates to an image and renders a model view of the surfel map using OpenGL and the currently estimated pose. This allows easy association between projected points and rendered surfels and enables frame-to-model alignment via \ac{ICP} with the point-to-plane metric. Afterwards, the map integrates the new scan surfels with an exponentially moving average if they are more accurate. A binary Bayesian Filter estimates the stability and reliability per surfel such that only stable surfels are kept. A pose graph reduces overall drift upon loop-closures and directly deforms the surfel map via the sensor poses.

Instead of a uniform resolution, MRSLaserMaps~\cite{david2017ras} represents the environment close to the sensor with higher detail. The registration performs \ac{EM} of the joint log-likelihood that a scan surfel is an observation of the \ac{gmm} of the local map. Circular buffers over grid cells and the fixed number of points stored within each cell enable map shifting to preserve the map's egocentric property. The shifted local map is added to a pose graph to reduce drift over time using constraints between neighboring local maps.
The approach was extended in \cite{david2018icra} to a hierarchical pose graph and a refinement step to realign scans within a previous local map. These scans were further undistorted by a least-squares fit of a cubic Lie group B-Spline to interpolate between  scan poses. In practice, this method provided offline generated maps for robot localization~\cite{beul2018iros}.

Elastic LiDAR Fusion~\cite{park2017icra} uses only a linear continuous-time trajectory. Here, a single transformation linearly interpolates the trajectory within a time segment under an inherent constant-velocity assumption. For the rotation, the Log map of the Lie group $SO(3)$ lifts the relative rotation between start and end pose to its vector-spaced Lie algebra $\mathfrak{so}(3)$, where the interpolation itself takes place. The exponential map Exp maps back to the interpolated rotation. This simple strategy is quite efficient and fast but is limited in practice by the constant-velocity assumption. Their trajectory optimization facilitates geometric constraints penalizing deviations along the normal direction between individual surfels at different time steps within the same scan as well as towards the global map and inertial constraints for rotational velocity and acceleration from IMU. Furthermore, the authors improve map consistency on loop closures through a deformation graph.

\begin{figure}[t]
  \centering
  \resizebox{1.0\linewidth}{!}{\begin{tikzpicture}
[content_node/.append style={font=\sffamily,minimum size=1.5em,minimum width=6em,draw,align=center,rounded corners,scale=0.65},
label_node/.append style={font=\sffamily,scale=0.5},
group_node/.append style={font=\sffamily,dotted,align=center,rounded corners,inner sep=1em,thick},>={Stealth[inset=0pt,length=4pt,angle'=45]}]

\definecolor{red}{rgb}     {0.5,0.0,0.0}
\definecolor{green}{rgb}   {0.0,0.5,0.0}
\definecolor{blue}{rgb}    {0.0,0.0,0.5}
\definecolor{grey}{rgb}    {0.5,0.5,0.5}

\draw[thick, rounded corners, grey!20!white,fill] (-1.8,2.2) -- (-1.8,3.8) -- (0.25,3.8) -- (0.25,2.2) -- cycle;
\node(Label)[label_node,align=center,anchor=south west] at (-1.6,3.8){\textbf{Sliding} \\ \textbf{Registration} \\ \textbf{Window}};

\draw[thick, rounded corners, grey!20!white,fill] (4.6,2.) -- (4.6,4.0) -- (7.1,4.) -- (7.1,2.) -- cycle;
\node(Label)[label_node,align=left,anchor=south west] at (5.4,4){\textbf{Local Map}};

\node(Reg)[content_node,fill=blue!15!white,minimum width=3cm,minimum height=1.5cm] at (2.5,3.0){\textbf{Registration}};

\node(LIDAR)[content_node,fill=green!15!white] at (-1,5) {LiDAR};

\node(Scan1)[content_node,fill=blue!15!white] at (-1,3.5) {Scan $t_l$};
\node(ScanN)[content_node,fill=blue!15!white] at (-1,2.5) {Scan $t_{l-n}$};

\node(KF1)[content_node,fill=blue!15!white] at (6.3,3.5) {Keyframe $1$};
\node(KFN)[content_node,fill=blue!15!white] at (6.3,2.5) {Keyframe $N$};

\node(Spline)[content_node,fill=blue!15!white,rotate=90,anchor=north] at (-0.2,3) {Spline};

\node(Map)[content_node,fill=blue!15!white,rotate=90,anchor=north] at (4.7,3) {Local Surfel Map};

\node(CheckKF)[content_node,fill=blue!15!white] at (-1,1.5) {Create\\Keyframe?};

\draw[->,thick] (LIDAR.180) -- (LIDAR.180 -| -2,0) -- node[label_node,midway,below,rotate=90] {\SI{10}{\hertz}} node[label_node,midway,above,rotate=90] {Scan} (-2,0 |- Scan1.180) -- (Scan1.180);

\draw[dotted,thick] (Scan1.270) -- (ScanN.90);
\draw[dotted,thick] (KF1.270) -- (KFN.90);
\draw[dashed,->,thick] (KF1.180) -- (Map.270);
\draw[dashed,->,thick] (KFN.180) -- (Map.270);

\draw[->,thick] (ScanN.270) -- node[label_node,midway,right] {\SI{10}{\hertz}} node[label_node,midway,left] {Scan, Pose} (CheckKF.90);

\draw[->,thick] (CheckKF) -- node[label_node,midway,below] {\SI{10}{\hertz}} node[label_node,midway,above] {Scan, Pose} (CheckKF.0-| KFN.270) -- (KFN.270);

\draw[->,thick] (Spline.270) -- node[label_node,midway,below] {Spline}
node[label_node,midway,above] {Surfels} (Reg.180);

\draw[->,thick] (Reg.270) -- (Reg.270 |- 0,2 ) -- node[label_node,midway,above] {Update}  (Spline.180 |- 0,2 )-- (Spline.180);

\draw[->,thick] (Map.90) -- node[label_node,midway,above] {Surfels} (Reg.0);

\end{tikzpicture}}
  \caption{System overview: 
A continuous-time trajectory spline describes each scans pose within the current sliding-registration-window. The registration aligns the scan surfels with a local surfel map and updates the spline. Keyframes are added if necessary to the sliding-window of the local map and combined in the local surfel map.}
  \label{fig:system}
\end{figure}
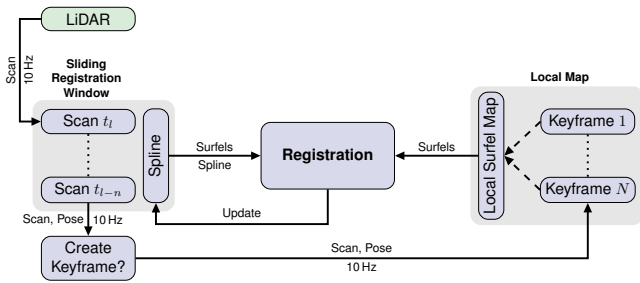

Although we base our work upon MRSLaserMaps, our real-time odometry system (\reffig{fig:system}) is a full redesign with robustness and efficiency in mind to cope with the large number of scan points generated by modern LiDAR sensors.
While most odometry systems align each LiDAR scan individually against the map, we jointly register multiple scans at once in a sliding-window fashion using the continuous-time trajectory B-Spline representation by Sommer \etal\cite{sommer2020cvpr}.
In contrast to MRSLaserMaps, we do not use dense but sparse voxel grids or lattices for each level within our multi-resolution surfel map. Additionally, we scale the surfels' \ac{gmm}  weight to balance the influence of differently sampled areas due to sensor geometry. We adaptively select the appropriate resolution for registration instead of the finest available. Furthermore, we fuse and shift maps via their surfels instead of point-wise and apply a keyframe-based sliding-window for the local map such that we integrate only scans with differing view poses.

\section{Our Method}
The point cloud $P$ of a scan is a set of points $\mathbf{p}_i\in \mathbb{R}^3$.
The statistics of all points falling into a single voxel cell are represented by a surfel with their mean value $\mathbf{\mu}$ and the covariance $\Sigma$. The surfel normal $\mathbf{n}$ is computed as the eigenvector to the smallest eigenvalue of $\Sigma$.
We regard a surfel as \textit{valid}, if it represents at least $10$ points and at least the two largest eigenvalues are non-zero.

\subsection{Multi-resolution Surfel Map}
Our map covers a cubic volume with side length $b$, \eg twice the sensor range. The map origin is at the center of this volume. 

We subdivide this volume in a uniform \ac{voxel} grid or a \num{3}-dimensional permutohedral lattice with tetrahedral cells. In both cases, we set the distance between adjacent vertices/corners to fixed-size $m$.

\begin{figure*}[t!]
  \centering
  \resizebox{1.0\linewidth}{!}{\input{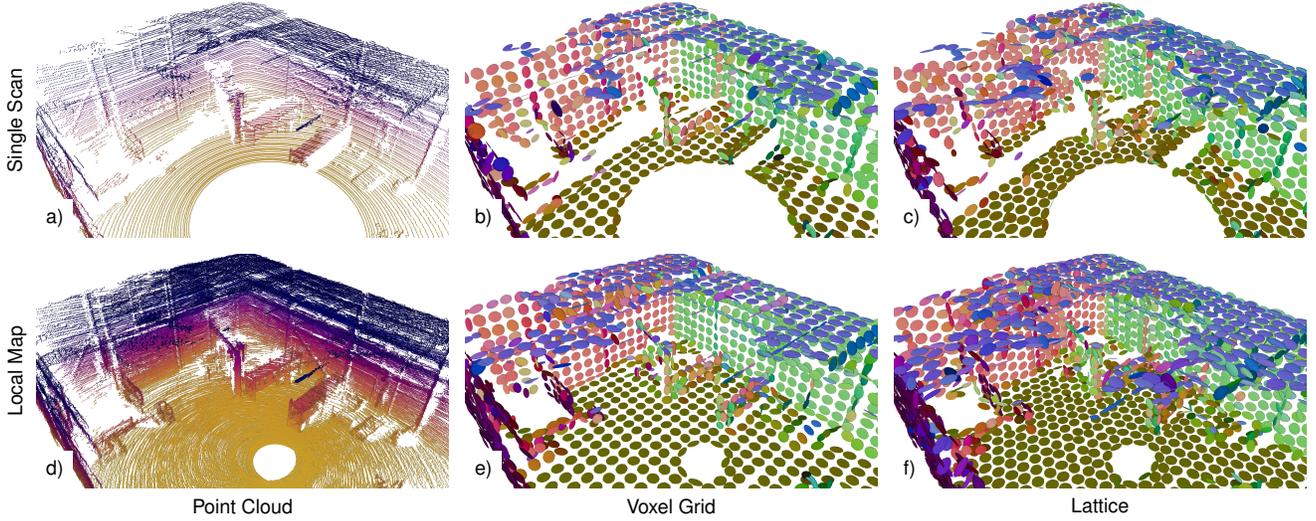}}
  \caption{Comparison between surfel maps with voxel grid [b),e)] and permutohedral lattice [c),f)].}
  \label{fig:lattice_grid}
\end{figure*}

A scaled regular grid $(d+1)\mathbb{Z}^{d+1}$ projected along the one vector $\mathbf{1}$ 
onto the hyperplane $H_d: \mathbf{p}\cdot \mathbf{1}=0$ forms the lattice~\cite{rosu2019latticenet}. Each vertex has $2(d+1)$ neighbors and its coordinates sum up to zero. In the lattice, we attribute all points to their closest vertex. A simple rounding algorithm provides the enclosing simplex for each point. One advantage of the lattice compared to the regular voxel grid is the better scalability for higher dimensions. More importantly, a vertex has $2\left(d+1\right)$ direct neighboring vertices instead of $3^d$, which is more efficient for our soft assignments.
Similarly, we do not allocate the whole dense volume but use a sparse lattice or a (block)-sparse voxel grid based on hash-maps.
For a more appropriate representation of the sensor geometry, we increase the detail close to the sensor by introducing multiple finer levels. With an increasing level, we half the side length, keep the number of cells constant, and centered at the origin. For registration, the finest available resolution is used or adaptively selected (\refsec{sec:adapt}).

Within a cell, we have a \ac{deque} to store individual surfels from different scans and a combined surfel. The \ac{deque} enables easier and faster removal of old scans from the map, in comparison to MRSLaserMaps. Adding a new scan creates a new surfel on top of the \ac{deque}. New scan points are fused only into the new surfel. The cell center is subtracted prior to fusion for numerical stability of the surfel. After processing all points, the changed cells update the combined surfel. During registration, we only use the combined surfels. \reffig{fig:lattice_grid} shows an example for a single scan as well as the local map with either a voxel grid or the lattice.

Each map stores its pose $T_{ws}\in SE(3)$ and the sensor origin $\mathbf{o}\in\mathbb{R}^3$, since these may change during map fusion. Every scan within the sliding registration window has its own surfel map, placed at its respective origin. The scan is added at the origin with identity transform. 
Although this results in differing grid alignment between the scan's surfel map and the local map, it greatly simplifies and speeds up the registration. Grid realignment of all scans within the window is costly. Using the current estimate for the newly added scan would require a more complex spline and would take longer to optimize. Furthermore, we alleviate the grid differences with the soft assignment (\refsec{sec:reg}).

We adopt a sliding-window keyframe approach for our local map. Once the distance towards the previous keyframe becomes too large, we add the last scan $P$ from the sliding registration window to the local map.
If there are too many keyframes in the local map, we remove the oldest one, including the  corresponding surfels from the cells \ac{deque} and recompute the combined surfel after adding the new scan.
The point cloud $P$ in the sensor frame is transformed to the local map frame and then integrated.
Before integration, we compute how many cells the new sensor pose moved away from the origin on the coarsest level. Once the difference is above a certain threshold, we shift the whole map to maintain its egocentric property. Shifting by a multiple of the coarsest cell enables efficient swapping of cells instead of cumbersome recomputation from points.

\subsection{Sliding-Window Continuous-time Trajectory Registration}
\label{sec:reg}
We represent our trajectory using the cumulative B-Spline formulation of Sommer \etal\cite{sommer2020cvpr} on $\mathbb{R}^3$ for the translation and on the Lie group $SO(3)$ for the rotation. A B-Spline of order $k$ itself is $C^{k-1}$ continuous and defined by its $k$ control points~\cite{sommer2020cvpr}:
\begin{align}
i\left(t\right) &= \lfloor(t-t_0) / \Delta t\rfloor, \\
u\left(t\right)&=((t-t_0) \bmod \Delta t) / \Delta t, \\
\mathbf{p}\left(u\right) &= \mathbf{p}_i \cdot \prod_{j=1}^{k-1}\lambda_j\left(u\right)\cdot \left(\mathbf{p}_{i+j}-\mathbf{p}_{i+j-1}\right),\label{eq:pos}\\
R\left(u\right) &= R_i \cdot \prod_{j=1}^{k-1}\text{Exp}\left(\lambda_j\left(u\right)\cdot \mathbf{d}^i_j\right),\label{eq:rot}\\
\mathbf{d}^i_j &= \text{Log}\left(R_{i+j-1}^{-1}R_{i+j}\right) \in \mathbb{R}^3,\label{eq:diff}\\
T_X\left(t\right) &= \left[\begin{matrix}R\left(u\left(t\right)\right) & \mathbf{p}\left(u\left(t\right)\right)\\ \mathbf{0} & 1\end{matrix}\right].
\end{align}
Here, $i$ is the index of the first control point in the active time segment $\left[t_i,t_{i+1}\right)$ and $u$ is the normalized time since the start of that segment.
For $R\in SO(3)$ Log maps from the Lie group to its vector-spaced Lie algebra and Exp maps back to $SO(3)$. Although the control points $X_i=\left(R_i,pi\right) \in SO(3)\times\mathbb{R}^3$ are from two different Splines, the result $T\left(t\right)$ is a rigid transform in $SE(3)$. 

In practice, we set the order to $k=3$ and thus, have to optimize all $k$ control points during registration --- independent of the number of scans $n$. The time interval $\Delta t$ is adaptively chosen such that the last scan $l$ in the sliding-window is within the current time interval $\left[t_{l-n},t_{l}+t_{pred}\right]$. Here, $t_{l-n}$ is the time of the last shifted out scan. We update the control points after shifting through the following optimization:
\begin{align}
X_{init} &= \argmin_X \sum_{i=1}^n \mathbf{e}_\mathbf{p}^\intercal \mathbf{e}_\mathbf{p} + \mathbf{e}_R^\intercal \mathbf{e}_R,\\
\mathbf{e}_\mathbf{p} &= \mathbf{p}\left(u\left(t_j\right)\right)-\mathbf{p}_{j-1},\\
\mathbf{e}_R &= Log\left(R_{j-1}^{-1}R\left(u\left(t_{j}\right)\right)\right),\\
j &= l-n+i,
\end{align}
where $R_{j-1}$ and $\mathbf{p}_{j-1}$ are the previous estimates for the unshifted spline.
The prediction time offset $t_{pred}$ helps during the deployment of the system on real robots to predict the robot's position, \eg \SI{0.1}{\second} or one scan into the future, effectively reducing the lag due to processing time.

We model the likelihood of a scene surfel $s$ observing a map surfel $m$ as the following normal distribution:
\begin{align}
\mathbf{e}_{sm}\left(T\right) &\sim \mathcal{N} \left( \mathbf{d}_{sm} , \Sigma_{sm} + \sigma^2I\right),\label{eq:obs_like}\\
\mathbf{d}_{sm}\left(T\right) &= T\mathbf{\mu}_s - \mathbf{\mu}_m,\\
\Sigma_{sm}\left(T\right) &= \Sigma_m + R\Sigma_sR^\intercal,
\end{align}
with $R$ being the rotation of the rigid transform $T$ and a resolution-depending scaling term $\sigma^2$. For better readability, we will drop the transformation argument for above formulas whenever possible.
This model would require a hard decision whether or not $s$ corresponds to $m$. Instead, we apply a \ac{gmm} that represents a mixture of multiple map surfels being observed by $s$ following the normal distribution \eqref{eq:obs_like} with an additional uniform component $p(o_s)$ for outliers: 
\begin{align}
p_s\left(T\right) &= p\left(o_s\right) + \sum_{m\in\mathcal{A}_s} p\left(a_{sm}\right) p\left(\delta_{sm}\right) p\left(\mathbf{e}_{sm}\right),\\
p\left(o_s\right)  &= p\left(o\right) p\left(\mathcal{N}\left(\mathbf{0},R\Sigma_sR^\intercal+\sigma^2I\right)\right).
\end{align}
Here, $\mathcal{A}_s$ is the set of map associations for surfel $s$ with the prior association likelihood $p(a_{sm})$. For more details on the \ac{gmm}, we refer the reader to \cite{david2017ras}.

$p(\delta_{sm})$ further takes the similarity between associated surfels \wrt normal and viewing direction into account. Both are modeled as: 
\begin{align}
p_v\left(\mathbf{v}\right) &\sim \mathcal{N} \left( \arccos\left( \left(R\mathbf{v}_s\right) \cdot \mathbf{v}_m \right),\left(\pi/8\right)^2 \right).
\end{align}
Additionally, we model the distance in the normal direction:
\begin{align}
d_n &\sim \mathcal{N}\left(\mathbf{n}_m^\intercal\Sigma_{sm}^{-1}\mathbf{d}_{sm},\sigma^2\right),
\end{align}
assuming indepedence results in the approximation:
\begin{align}
p(\delta_{sm}) &= p_v(\mathbf{n}) p_v(\mathbf{v}) p(d_n).
\end{align}

Given this \ac{gmm} formulation, we seek to find the spline control points $X$ that maximize the logarithm of the joint observation likelihood over the mixture:
\begin{align}
X^\star = \argmax_X \sum\limits_{i=1}^{n} \sum_{s\in \mathcal{S}} \log\left(p_s\left(T_X\left(t_{l-n+i}\right)\right)\right),\label{eq:joint_likelihood}
\end{align}
and solve \eqref{eq:joint_likelihood} with \ac{EM}. In the E-Step, we establish associations for all surfels within the sliding-window. The pose $T_c$, evaluated the spline at time $t_c$, transforms the mean $\mu_s$ from the scan's sensor frame into the map frame. Then, we lookup the corresponding surfel in the map and check its 1-hop-neighborhood for valid surfels. In 3D, this creates up to \num{27} scan-map associations for a 3D voxel grid or \num{9} for the lattice.
We calculate per association their conditional likelihood $w_{sm}$ given the current estimate $T_c$:
\begin{align}
w_{sm} \left(T_c\right) &= \frac{p\left(a_{sm}\right) p\left(d_{sm}\right) p\left(\mathbf{e}_{sm}\right)}{p_s(T_c)},
\end{align}
and keep these associations and weights fixed during the M-Step to compute an updated estimate for the control points $X$:
\begin{align}
X^\star &= \argmin_X \sum\limits_{i=1}^{n} \sum_{s\in \mathcal{S}_i} \sum_{m\in\mathcal{A}_s} r_{sm}(T_X(t_{l-n+i})),\label{eq:m_step}\\
r_{sm}(T_c) &= w \, \mathbf{d}^\intercal \Sigma^{-1} \mathbf{d}.
\end{align}
The Levenberg-Marquardt algorithm optimizes \eqref{eq:m_step}.

We found empirically that the \ac{gmm} assigns a higher weight $w_{sm}$ to surfels in the 3D-LiDAR's vicinity. This leads in situations where one translation direction was underconstrained, e.g. open park areas with tree trunks further away, or dynamic environments with moving objects close to the sensor, to wrongly estimate the translation as staying in place while the orientation was correct. Thus, following HeRO~\cite{hero}, we analyzed the condition number $\kappa$ of the covariance of surfel normals --- since $\kappa$ reflects the difficulty to accurately solve a linear system of equations. Weighting the normals with their conditional likelihood $w_{sm}$ increases $\kappa$ --- in some cases by a factor of up to \num{50}. We found that this relates to the higher measurement density for surfaces close to the LiDAR, because the prior association likelihood $p(a_{sm})$ directly incorporates the number of measurements per surfel and the LiDAR samples the environment non-uniformly in 3D. Hence, we inversely weight $w_{sm}$ by the number of surfel measurements to effectively level the influence between far and close surfels.

\subsection{Adaptive Resolution Selection}
\label{sec:adapt}
Always restricting to the finest available map resolution for registration is computationally inefficient for planar surfaces like walls, roads, floors, or ceilings.
Hence, we select the most appropriate resolution adaptively.

\begin{figure*}
  \centering
  \resizebox{1.0\linewidth}{!}{\input{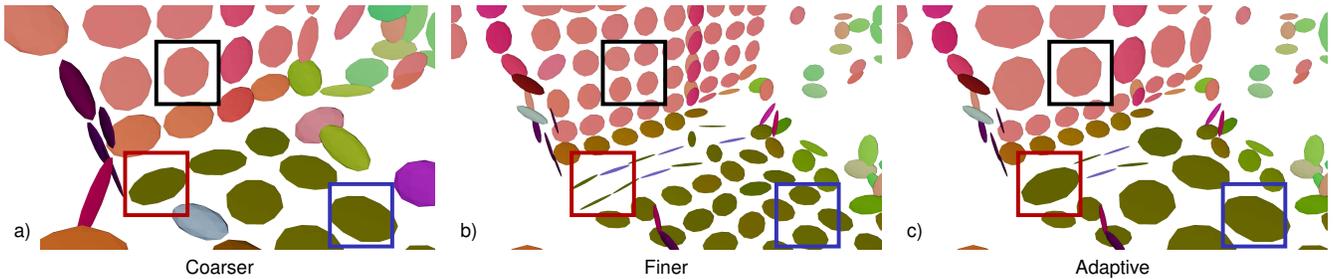}}
  \caption{Adaptive resolution selection selects coarser surfel in areas of finer resolution with planar surfels. The red rectangle highlights merging multiple degenerate surfels, while the black and blue are examples for \refeq{eq:Sel1} and \refeq{eq:Sel2}.}
  \label{fig:adaptive_selection}
\end{figure*}

We start collecting valid surfels on the finest map scale and check for not yet processed valid surfels for the following three conditions on the normalized Eigenvalues $\lambda_0 \leq \lambda_1 \leq \lambda_2 \in \mathbb{R}\, s.t. \sum_i \lambda_i = 1$ of the surfel covariance matrix $C$:
\begin{align}
\lambda_0 &< \theta_{planar},\label{eq:Sel1}\\
\lambda_0 &< \lambda_1\theta_{scale},\label{eq:Sel2}\\
\lambda_1 &< \theta_{degenerate}.
\end{align}
The first two cases directly relate to planar surfaces, while the last case often occurs for surfels created from a single scan line. \reffig{fig:adaptive_selection} illustrates these.
If at least one condition is true, we test all valid neighbors falling into the same coarser surfel and calculate the mean normal vector $\bar{\mathbf{n}}$.
Once all valid finer surfels pass this test, we perform the same check on the coarser surfel. Additionally, we ensure that the mean and coarser normal are similar:
\begin{align}
\abs{\mathbf{n}_c \cdot \bar{\mathbf{n}}} > \theta_n.
\end{align} Upon passing these tests, the coarser surfel will be further processed instead of the smaller ones, thus, reducing the total number of surfels and speeding up registration. Typical values for the thresholds are $\theta_{planar} = 0.01$, $\theta_{scale}=0.01$, $\theta_{degenerate}=0.1$, $\theta_n=0.8$.

\section{Evaluation}
We evaluate our method on the Urban Loco dataset~\cite{wen2020icra} for autonomous driving, 
the Newer College dataset~\cite{ramezani2020newer}, and nine self-recorded UAV flights through the DRZ Living Lab.
Furthermore, we show qualitative results from multiple UAV flights.
All experiments were conducted on a laptop with an Intel Core i7-6700HQ with \SI{32}{\giga\byte} of RAM and an NVIDIA GeForce GTX 960M.
We compare our method against the two LOAM variants A-LOAM\footnote{\url{https://github.com/HKUST-Aerial-Robotics/A-LOAM}} and F-LOAM\footnote{\url{https://github.com/wh200720041/floam}} as well as SuMa\footnote{\url{https://github.com/jbehley/SuMa}}. We modified A/F-LOAM to use ROS \textit{message\_filters}\footnote{\url{http://wiki.ros.org/message_filters}} to enforce real-time with a limited queue size of 10 to prevent processing too old information. Furthermore, we deactivated for fair comparison the loop-closing of SuMa. We did not include MRSLaserMaps since it lost track on all tested sequences.
Our method is implemented in C++ and runs on the CPU only. In all experiments we perform up to three iterations within the Levenberg-Marquardt algorithm to optimize \eqref{eq:m_step}. The storage type is by default set to a sparse voxel grid unless otherwise noted.

For all systems, we evaluate the \ac{ATE} for the at-runtime estimated poses against the provided ground-truth. The best result will be in bold, while the second best will be colored blue. For both LOAM-derivatives, we report the more accurate optimized poses instead of the initial odometry poses and ignore unoptimized scans during \ac{ATE} computation.

\subsection{Newer College Dataset}
Ramezani \etal\cite{ramezani2020newer} replicated the route of the New College Dataset with a handheld Ouster OS-1 LiDAR with 64 beams at \SI{10}{\hertz}. We selected the sequences ``01\_short`` and ``02\_long`` and ``05\_quad\_with\_dynamics``.
The first two experiment runs ``01`` and ``02`` consist of multiple loops between buildings and through a park for over \SI{1530}{\second} and \SI{2656}{\second}. Sequence ``05\_quad\_with\_dynamics`` is a shorter \SI{398}{\second} run with faster and partially swinging motion during four loops within a quad. The authors aligned the LiDAR scans against a terrestrial laser scanner to provide ground-truth poses. 
\reftab{tab:newer_college} shows the resulting ATE. Although the LOAM-derivates' odometry processed each scan, both could only fully optimize every third scan in real-time.  
On the second sequence, SuMa unfortunately lost track in the park area while A-LOAM and F-LOAM exhibited drift during orientation changes.
For sequence ``05`` the limited size of the quad and A-LOAMs' local map optimization allows A-LOAM to relocalize, thus keeping the error low. In contrast, our sliding-keyframe-window accumulates more drift over time but maintains high accuracy such that in the future the extraction of loop-closure candidates based on some vicinity criterion is possible.

\subsection{Urban Loco Dataset}
Wen \etal\cite{wen2020icra} equipped two cars with LiDAR and other sensors to capture highly urbanized areas throughout San Francisco and Hong Kong. A navigation system with RTK-GPS and IMU provides the ground-truth for all sequences.
The Coli Tower sequence (ULCT) took \SI{248}{\second} and is a \SI{1.8}{\kilo\metre} long drive up hill within a dynamic environment. Similarly, the Lombard Street sequence (ULLS) is a \SI{1}{\kilo\metre} drive and took \SI{253}{\second}. Both were captured with a RS-LiDAR-32 while the seven Hong Kong sequences used a Velodyne HDL-32E LiDAR. The Hong Kong sequences took between \SI{150}{\second} and \SI{365}{\second} and are up to \SI{2}{\kilo\metre} long through dense urban environment.

\reftab{tab:newer_college} shows the results for all four algorithms. Our method achieves consistent state-of-the-art results with placing first in three and second in another two out of nine sequences.

\bgroup
\newcolumntype{Y}{>{\centering\arraybackslash}X}
\renewcommand{\arraystretch}{1.3}  
\begin{table}
\centering
\caption{ATE~[\si{\metre}] Evaluation on scenes from the Newer College (NC) and Urban Loco (UL) datasets}
\begin{tabularx}{\columnwidth}{
l|Y|Y|Y|Y|Y|}
\cline{2-6}
 & Ours (Grid) & Ours (Lattice) & A-LOAM & F-LOAM & SuMa \\\hline
NC01 & 2.17229 & $\bm{1.97836}$ & 3.30768 & 101.899 & \tcb{2.04814} \\\hline
NC02 & $\bm{4.93553}$ & \tcb{5.11236} &  62.64240 & 87.08770 & X \\\hline
NC05 & 0.46826 & \tcb{0.41505} & $\bm{0.14824}$ & 2.81843 & 1.87838 \\\hline
ULCT & \tcb{4.99616} & $\bm{4.96610}$ & 10.37230 & 6.05617 & 7.77146 \\\hline 
ULLS & 7.82386 & \tcb{7.57318} & 8.40002 & $\bm{7.56019}$ & 10.69550 \\\hline 
ULHH & \tcb{2.35564} & 2.40788 & 2.43283 & 2.42415 & $\bm{2.32560}$ \\\hline
ULLL & 2.73179 & \tcb{2.23304} & $\bm{2.07094}$ & 2.31379 & 8.44114 \\\hline 
ULSL & 3.19269 & 3.26606 & \tcb{3.16574} & 3.27051 & $\bm{2.87772}$ \\\hline 
ULWH & 2.95064 & 2.99434 & \tcb{2.42352} & $\bm{2.25923}$ & 3.50044 \\\hline
ULT2 & 2.54422 & $\bm{1.44114}$ & 17.34920 & 2.42821 & \tcb{2.21693} \\\hline
ULT3 & \tcb{1.59692} & $\bm{1.55525}$ & 9.67000 & 2.02969 & 2.94482 \\\hline
ULH5 & \tcb{1.55182} & $\bm{1.50516}$ & 17.35210 & 2.42821 & 2.21693 \\\hline
\end{tabularx}
\label{tab:newer_college}
\end{table}
\egroup

\subsection{DRZ Living Lab}
We collected LiDAR scans from an Ouster OS-0 with 128 beams at \SI{10}{\hertz} attached to a DJI M210 v2 while flying through the DRZ Living Lab\footnote{\url{https://rettungsrobotik.de/living-lab/}}. A \ac{mocap} system provides ground-truth poses in the starting section of the lab, where we recorded multiple flights with up to \SI{2.5}{\minute} flight time. All sequences except for the second dataset remain within the \ac{mocap} volume with varying angular velocity and linear acceleration. In the second dataset we traversed back and forth through the Living Lab. The flight starts and finishes within the \ac{mocap}. Three further experiments were conducted with three persons moving around with a slow (3P-S), medium (3P-M) and fast (3P-F) flying UAV. For comparison, we conducted similar exploration flights with different motion speeds in a static environment. For fair evaluation, we use the UAVs' IMU orientation to compensate for the rotation distortion and supply these scans to all tested methods.
\reftab{tab:drz_halle} details the number of scans, the median norm of the linear acceleration as well as the median and maximum norm of the angular velocity and shows the results for all algorithms.

The surfel-based methods cope well with the faster moving sequences, while the LOAM-derivatives struggle, even with the undistorted scans. 
Our method performs very well in comparison, with the exception of the ``hall``-sequence. 
This has likely to do with the local map size of all methods, since our sliding-keyframe-window is the only one that did not cover the whole Living Lab.

To analyze the impact of different spline settings, we evaluate multiple combinations on the most challenging sequence ``F2``. \reftab{tab:stats} reports the parameters for number of control points $k$, the number of jointly optimized scans $n$ and the resulting \ac{ATE} with the actual computation time during registration. The storage type was set to sparse grid since timings for block-sparse and sparse were similar due to the relatively low surfel count. Increasing $n$ provided no benefit, but slows down computation and required in some cases more iterations to compute a stable result.
In comparison, the registration took \SI{29.82}{\milli\second} when using a permutohedral grid with $k=3$ and $n=3$ without adaptive resolution selection at an \ac{ATE} of \SI{0.09489}{\metre}. The resolution selection reduced runtime to an avg. \SI{22.95}{\milli\second}, although the ATE increased slightly to \SI{0.10016}{\metre}.
In contrast, the voxel grid without adaptive selection could not correctly estimate all rotations resulting in a \SI{90}{\degree} drift during a high angular velocity maneuver.
Similarly, we typically obtained more consistent and accurate results with enabled adaptive resolution selection in open environments.

\bgroup
\newcolumntype{Y}{>{\centering\arraybackslash}X}
\renewcommand{\arraystretch}{1.3}  
\begin{table*}
\centering
\caption{ATE~[\si{\metre}] Evaluation for the DRZ Living Lab dataset}
\begin{tabularx}{\linewidth}{
l|c|Y|Y|Y|Y|Y|Y|Y|Y|}
\cline{2-10}
Seq. & Scans & $\med{\norm{\bm{a}}}$ [\si{\metre\per\square\second}] & $\med{\norm{\bm{\omega}}}$ [\si{\radian\per\second}] & $\max(\norm{\bm{\omega}})$ [\si{\radian\per\second}] & Ours (Grid) & Ours (Lattice) & A-LOAM & F-LOAM & SuMa \\\hline
Fast & 1479 & 1.13119 & 0.22876 & 1.89359 & 0.094848 & $\bm{0.05566}$ & 2.26457 & 2.11024 & \tcb{0.06370} \\\hline 
Hall & 2421 & 0.39893 & 0.09758 & 1.14679 & 0.20538 & 0.09212 & $\bm{0.03032}$ & \tcb{0.03850} & 0.07256 \\\hline 
3P-S & 1085 & 0.27435 & 0.07900 & 0.86189 & 0.02181 & $\bm{0.01465}$ & 0.02244 & \tcb{0.01665} & 0.03834 \\\hline 
3P-M &  650 & 0.58921 & 0.11351 & 1.08112 & 0.02873 & $\bm{0.02393}$ & \tcb{0.02517} & 0.03076 & 0.04032 \\\hline 
3P-F &  604 & 1.99022 & 0.52543 & 3.54729 & \tcb{0.06066} & $\bm{0.04995}$ & 0.71797 & 0.83169 & 0.09472 \\\hline 
S1   &  826 & 0.48876 & 0.10443 & 1.51561 & 0.04398 & $\bm{0.03968}$ & 0.04555 & 0.04636 & \tcb{0.041957} \\\hline 
M1   & 1458 & 0.90011 & 0.21923 & 1.97838 & \tcb{0.08220} & $\bm{0.06385}$ & 3.73529 & 3.62154 & 0.17566 \\\hline 
F2   &  795 & 2.30749 & 0.48656 & 3.98107 & 0.10458 & \tcb{0.10016} & 2.62554 & 2.80373 & $\bm{0.08839}$ \\\hline 
F3   &  957 & 1.04914 & 0.24944 & 1.63187 & \tcb{0.08520} & $\bm{0.07374}$ & 0.14406 & 2.45972 & 0.17991 \\\hline 
\end{tabularx}
\label{tab:drz_halle}
\end{table*}
\egroup

\bgroup
\newcolumntype{Y}{>{\centering\arraybackslash}X}
\renewcommand{\arraystretch}{1.3}  
\begin{table}
\centering
\caption{Statistics for varying spline parameters on the ``F2`` sequence in the DRZ Living Lab. Entries with * required five instead of three iterations.}
\begin{tabularx}{\columnwidth}{
|Y|Y|Y|Y|}
\multicolumn{2}{c}{Spline} & ATE & avg. time\\
$k$ & $n$ & [\si{\metre}] & [\si{\milli\second}] \\\hline
\multirow{3}{*}{2} & 2 & 0.10628 & 41.28 \\\cline{2-4}
 & 3 & 0.10970* & 65.45 \\\cline{2-4}
 & 4 & 0.22468* & 71.03 \\\hline
\multirow{3}{*}{3} & 3 & 0.10458 & 50.57 \\\cline{2-4}
 & 4 & 0.10759* & 82.53 \\\cline{2-4}
 & 5 & 0.11229* & 91.85 \\\hline
\multirow{3}{*}{4} & 4 & 0.11133 & 117.03 \\\cline{2-4}
 & 5 & 0.11246 & 121.88 \\\cline{2-4}
 & 6 & 0.12333 & 121.88 \\\hline
\end{tabularx}
\label{tab:stats}
\end{table}
\egroup

\begin{figure*}
  \centering
  \resizebox{1.0\linewidth}{!}{\input{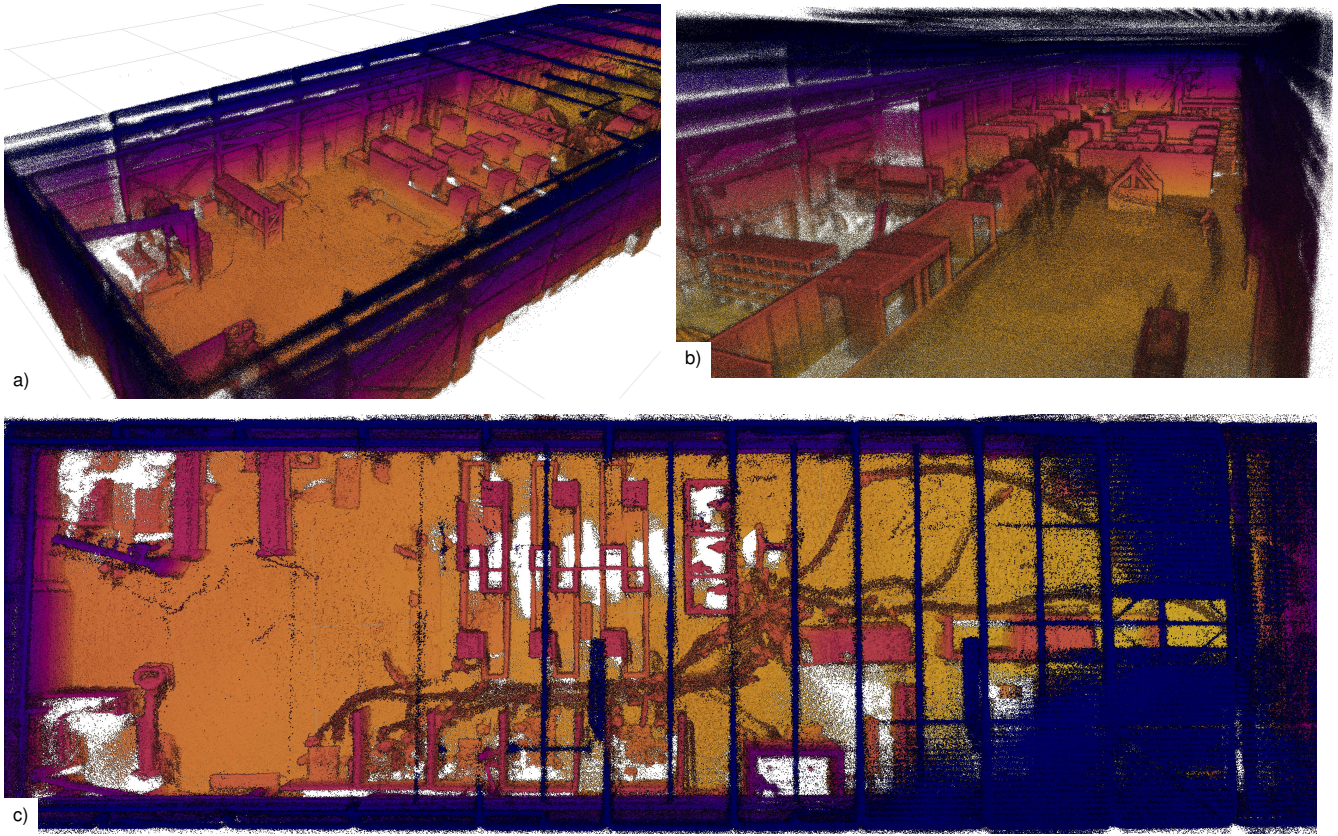}}
  \caption{Aggregated point cloud from the traverse through the DRZ Living Lab. The roof is partially removed for better visualization.}
  \label{fig:assembled_map_drz}
\end{figure*}

\reffig{fig:assembled_map_drz} shows the aggregated point cloud from the DRZ Living Lab traverse flight.

\subsection{Qualitative UAV Experiment}
\begin{figure*}
  \centering
  \resizebox{1.0\linewidth}{!}{\input{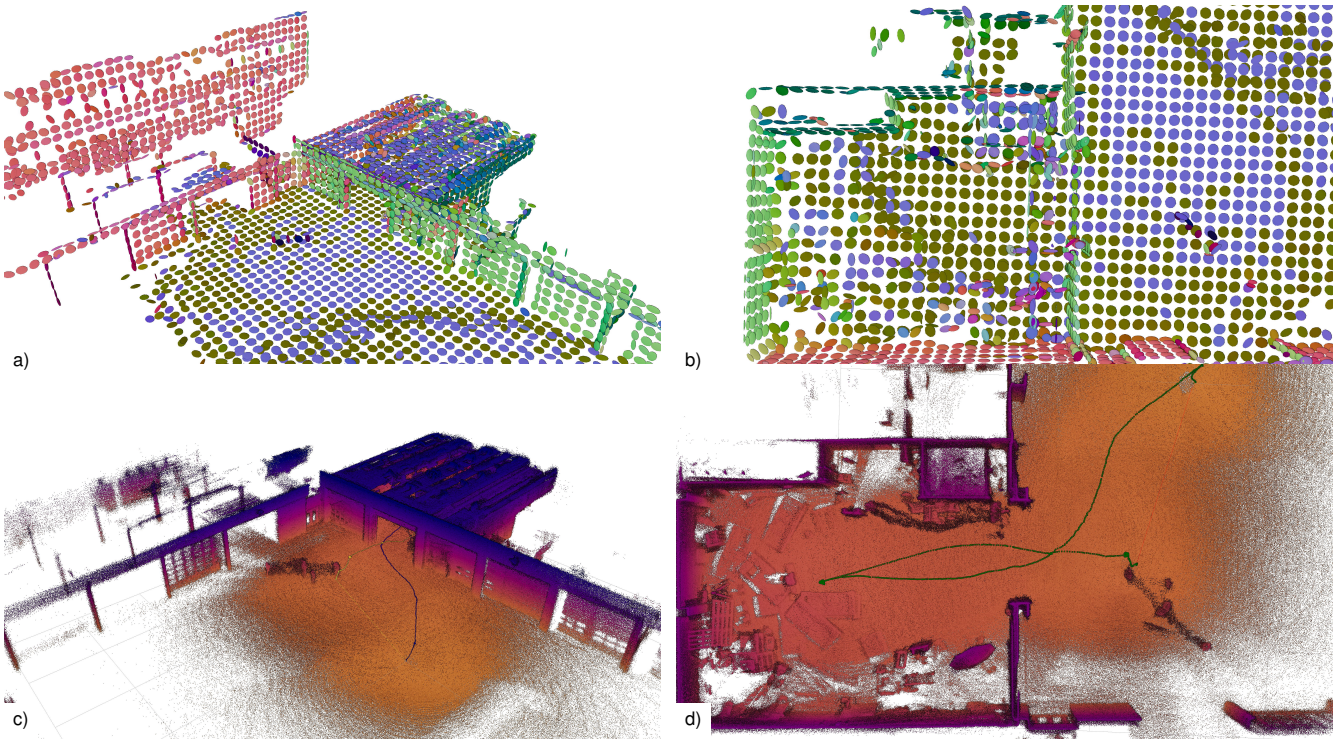}}
  \caption{Local map [a),b)] and aggregated point cloud [c),d)] and  from the autonomous flight. The trajectory is colored green in c) and d).}
  \label{fig:assembled_map_lbh}
\end{figure*}

An early development version of our registration provided the onboard LiDAR odometry for multiple autonomous flights of the same DJI M210v2 UAV through GNSS-denied areas. An EKF fused our pose estimates with the aircraft’s IMU. For more details on the autonomous UAV flights, we refer the reader to \cite{daniel2021icuas}\footnote{\url{https://ais.uni-bonn.de/videos/icuas2021_schleich/}}. \reffig{fig:assembled_map_lbh} shows the finest resolution of the local surfel map and the aggregated point cloud.

\section{Conclusion}
We presented MARS, a novel multi-resolution surfel-based LiDAR odometry. A sparse permutohedral lattice or voxel grid stores the surfels within our multi-resolution surfel map. Multiple LiDAR scans are jointly registered against the local map using a continuous-time trajectory with adaptively selected surfel resolution. In the future, we plan to incorporate IMU measurements and visual features in the registration process and extend our system with a pose graph to further reduce the drift over time, incorporating GPS poses and loop closures.


\bibliographystyle{IEEEtran}
\bibliography{literature}
\vfill
\balance

\end{document}